\documentclass[sigconf]{acmart}

\usepackage[algoruled,boxed,lined]{algorithm2e}
\usepackage{grffile} 
\usepackage{xcolor}
\usepackage{supertabular} 
\usepackage{cleveref}
\usepackage{multirow}
\usepackage{listings}

\usepackage{tabularx}

\usepackage{newfloat}
\DeclareFloatingEnvironment[
   fileext=lol,
   name=Listing
]{listing}

\usepackage{subcaption}
\DeclareCaptionSubType{listing}

\usepackage{mathtools} 
\usepackage{enumitem}

\acmConference[ICAIF '20]{ICAIF '20:
First ACM International Conference on AI in Finance}
{October 15, 2020}{New York}

\copyrightyear{2020}
\acmYear{2020}
\setcopyright{acmlicensed}\acmConference[ICAIF '20]{ACM International Conference on AI in Finance}{October 15--16, 2020}{New York, NY, USA}
\acmBooktitle{ACM International Conference on AI in Finance (ICAIF '20), October 15--16, 2020, New York, NY, USA}
\acmPrice{15.00}
\acmDOI{10.1145/3383455.3422563}
\acmISBN{978-1-4503-7584-9/20/10}
\begin{document}

\title{Deep Q-Network-based Adaptive Alert Threshold Selection Policy for Payment Fraud Systems in Retail Banking}

\author{Hongda Shen}
\affiliation{
  \institution{University of Alabama in Huntsville}
}
\email{hs0017@alumni.uah.edu}

\author{Eren Kurshan}
\affiliation{
  \institution{Columbia University}
}
\email{ek2529@columbia.edu}

\begin{abstract}
Machine learning models have widely been used in fraud detection systems. Most of the research and development efforts have been concentrated on improving the performance of the fraud scoring models. Yet, the downstream fraud alert systems still have limited to no model adoption and rely on manual steps. Alert systems are pervasively used across all payment channels in retail banking and play an important role in the overall fraud detection process. Current fraud detection systems end up with large numbers of dropped alerts due to their inability to account for the alert processing capacity. Ideally, alert threshold selection enables the system to maximize the fraud detection while balancing the upstream fraud scores and the available bandwidth of the alert processing teams. However, in practice, fixed thresholds that are used for their simplicity do not have this ability.
In this paper, we propose an enhanced threshold selection policy for fraud alert systems. The proposed approach formulates the threshold selection as a sequential decision making problem and uses Deep Q-Network based reinforcement learning. Experimental results show that this adaptive approach outperforms the current static solutions by reducing the fraud losses as well as improving the operational efficiency of the alert system.
\end{abstract}

\settopmatter{printfolios=true} 
\maketitle

\section{Introduction}
\label{sec:intro}

AI and machine learning models have extensively been used in payment fraud detection systems \cite{Turing19,Heaton16,WEF20} since \textit{1990s} \cite{Chan98, Bolton01,Ghosh94,Aleskerov97}. In recent years, fraud detection has been facing serious challenges due to the unprecedented growth in the digital payments. 
Global payment fraud losses are estimated to be well over \textit{25 Billion} U.S.D. per year \cite{Nilson19}\cite{Report19}. 
In response, financial institutions have made significant investments in upgrading and building in-house AI models. Supervised \cite{MarfaingG2018,NiuWY2019}, semi-supervised, unsupervised learning \cite{Bolton01}, deep neural networks \cite{Kang16}, decision trees \cite{Sahin13} and hybrid approaches \cite{Yan18} have broadly been explored towards deployment in the fraud detection systems. However, the primary (and for the most part only) focus of the AI and machine learning applications in fraud detection has been transaction fraud scoring. Meanwhile, many downstream systems still have limited AI modeling and rely on rule-based or manual processing stages.

Among these, \textit{Alert Processing} is a critical stage in fraud detection as it ties directly to suspicious and fraudulent transactions. Alerting systems are pervasively used in retail banking across all payment processing channels, including but not limited to credit card, debit card, ATM, person-to-person (P2P), check and deposit transactions, bill payment, online payment, wire and automated clearing house transactions (ACH) etc. Their challenges affect almost all channels and transaction types in retail banking.

One of the key parameters in an alert system is the \textit{alert threshold}, which is used to decide on which transactions to alert, based on their fraud scores and the available alert processing capacity.  Current alert systems use fixed thresholds derived from historical data. Due to their inability to adapt to the dynamic changes in the payment data these static thresholds frequently experience performance and operational issues.  Threshold selection is tied to some of the key aspects of the fraud detection systems:
\begin{itemize}[leftmargin=1em]
\item \textit{Fraud Detection Performance \& Fraud Losses:} Alert processing is directly linked to all suspicious or fraud transactions and their fraud losses.
\item\textit{Operational Costs \& Efficiency:}
Threshold selection is tied to the operational cost, performance and efficiency challenges. Static threshold based systems go through frequent periods of inefficiency during which (i) the alert processing teams do not have all relevant fraud cases in the alert population or (ii) true fraud alerts are dropped as high number of alerts cause alert processing capacity to be exceeded. 
\item \textit{Customer Satisfaction:}
In both scenarios, the failure to notify customers of true fraud cases or alerting non-fraud transactions create significant customer satisfaction challenges.
\end{itemize}
In this paper, we propose an \textit{Adaptive Threshold Selection Technique} for fraud alert generation. The proposed approach tries to improve the overall fraud detection performance based on the upstream AI model scores and potential fraud losses, while staying within daily alert processing capacity constraints. We formulate this into a sequential decision making problem and present a Deep Q-Learning based approach. Experimental results show that the proposed approach provides reduced fraud losses and improved customer satisfaction metrics.

\section{Fraud Detection in Payment Systems}

\begin{figure}[ht!]
\centering
\includegraphics[keepaspectratio,width=1\columnwidth]{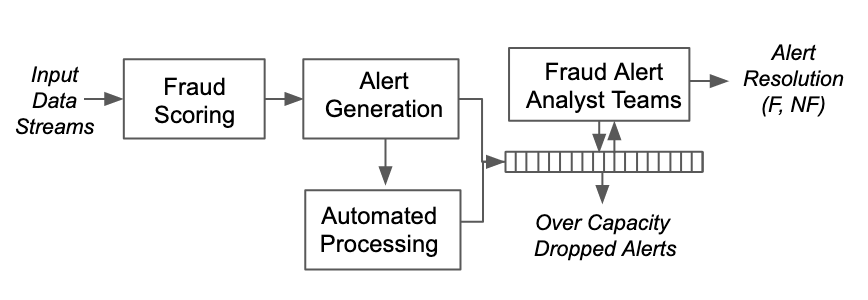}
\setlength{\abovecaptionskip}{-3pt}
\setlength{\belowcaptionskip}{-8pt}
\caption{High-level view of the fraud detection process flow}
\label{fig:workflow}
\end{figure}

Figure \ref{fig:workflow} illustrates a high-level view of the payment fraud detection process flow in financial services. 


\textit{AI/Machine Learning Model Scoring:}
AI and machine learning models are commonly used to score incoming payment transactions for fraud risk. The resulting model score is a key parameter in the downstream decision making processes.

\textit{Business and Fraud Strategy Rules:}
Following the model scoring, business and fraud strategies are applied as a rule engine. Business strategies focus on the current and downstream business impact of the transaction decision, while fraud strategies emphasize the broader risk by incorporating the fraud score generated by the machine learning model. 

\textit{Alert Generation and Resolution:}
At this stage, if the requirements are met, the system generates a \textit{fraud alert} and sends it to the customer. 
Alert processing is considered to be one of the most challenging stages in fraud detection systems.  By definition, alerted transactions have suspicious and non-confirming characteristics as opposed to the clearer separation of fraud and non-fraud in the overall transaction population. Fraud specialists utilize a multitude of internal and external data sources, channels and historical patterns for their investigations. They may contact the account owners to inquire about the transaction. As a result, the process is typically time consuming, complex and costly, which highlights the need for targeted optimization solutions.
 
Although most of the alert volume is processed through mobile applications/messaging in the state-of-the-art systems,it is not possible to fully eliminate manual alert processing. Some transactions are not eligible for the automated path and require analyst based manual processing. Similarly, when the customer fails to respond to alerts, current and downstream alerts are sent to the fraud alert specialists for investigations. Any alert that was not resolved by alert specialist teams due to bandwidth limitations is dropped or sometimes sent to backlogs.

\subsection {Threshold Selection}
As discussed in the Section \ref{sec:intro}, alert threshold selection ideally involves balancing multiple parameters in the system.
Static alert thresholds are not capable of optimizing for the alert processing capacity of the system. As their values remain fixed over time, they frequently experience periods of sub-optimal performance:
During the periods when the static alert threshold is too low for the system state, more customers than needed receive alerts for non-fraud transactions, degrading the customer satisfaction metrics. Frequently, the processing capacity is exceeded, and actual fraud alerts get dropped. In practice, dropped alert percentages can be quite high (e.g. well over 30\% of the capacity during over-alerting periods). 

On the other hand, when the static alert threshold is too high for the system state, the system misses more fraud cases by not alerting them, which in turn increases fraud losses. Customer satisfaction metrics degrade as the system fails to alert actual fraud cases.

As discussed earlier, dynamic capacity for the alert processing system is an important factor in the overall performance. Due to numerous factors (including environment changes, upstream rule and strategy changes, processing bandwidth fluctuations etc) each alert may experience a different processing capacity in real-time. This changes the available system capacity for a period of time such as one day or one hour. 

\section{Related Work}
Although there is an extensive amount of work on the AI and machine learning applications in fraud scoring, almost no research has focused on the alert processing systems or the alert threshold selection problem \cite{Survey_Tubb, Survey_Sorur}, \cite{Rule_Alert}. To the best of our knowledge, this is the first study to address the unique challenges of alert generation and processing through machine learning. Adaptive threshold selection provides solutions for the static threshold problems. However, in a dynamic environment, the ground truth for the actions taken cannot be predetermined and integrated in advance. Hence, building an adaptive threshold selection policy using conventional classification methods is not feasible. 

In this study, we use a Deep Q-Learning based approach \cite{MnihKSGAWR13,MinhKSR2015,HasseltGS2015}. Deep Q-Learning was initially proposed as a novel deep learning model for reinforcement learning to master control policies for Atari 2600 computer games. Later, it was adopted by a wide range of application domains such as resource management \cite{MaoAMK2016}, traffic light control \cite{ArelLUK2010},  personalized recommendation systems \cite{ZhengZZX2018,ChenLLJ2018}, and robotics \cite{LevineFDA2015,GuHLS2017}. In financial services, reinforcement learning has been proposed for market-making, portfolio optimizations, equities trading etc. \cite{XiongLZY2018,CharpentierER2020}.  In fraud detection, \cite{BouchtiCAO2017} used reinforcement learning for fraud attack design. We believe this study is the first use of reinforcement learning in fraud alert generation and fraud system optimization in financial services. 

\section{Proposed Approach}
In this section, we present a formulation of the fraud alert system optimization as a sequential decision making problem. Following the formulation, we provide an overview of the proposed Deep Q-Learning based technique.

\subsection{Problem Formulation}
The primary objective of the alert generation problem is to find the optimal score threshold to decide which transactions should be alerted. In practice, it is difficult to guarantee that an alerting decision is correct before the entire process concludes and the transaction is in fact deemed a fraud.  \textit{In other words, it is not possible to determine an optimal static threshold at decision time.} 
Static thresholds are typically calculated during scoring model development, based on the performance of a range of thresholds on the validation data set. 

As discussed in the Section \ref{sec:intro}, static thresholds gradually degrade in accuracy over the use period. The degradation is prominent in dynamic environments such as fraud detection. Static thresholds frequently cause overshoot and undershoot in the alert volume, cause high false positive rates, reduced customer satisfaction and missed fraud cases. In this paper, we use an adaptive threshold selection policy to replace the static approach. An hourly update period was chosen to demonstrate the approach through experimental analysis. Depending on the underlying application characteristics different time periods may be selected, which is beyond the scope of this work. The daily alert processing capacity is assumed to be $C_{max}$, i.e. the alert processing teams are processing alerts at full capacity which is a fixed value. Any alert above this value will be dropped. Due to the lack of alert resolution in dropped alerts, the fraud cases are not correctly labeled as fraud.

\begin{figure}[ht!]
\vspace{-5pt}
\centering
\includegraphics[keepaspectratio,width=0.85\columnwidth]{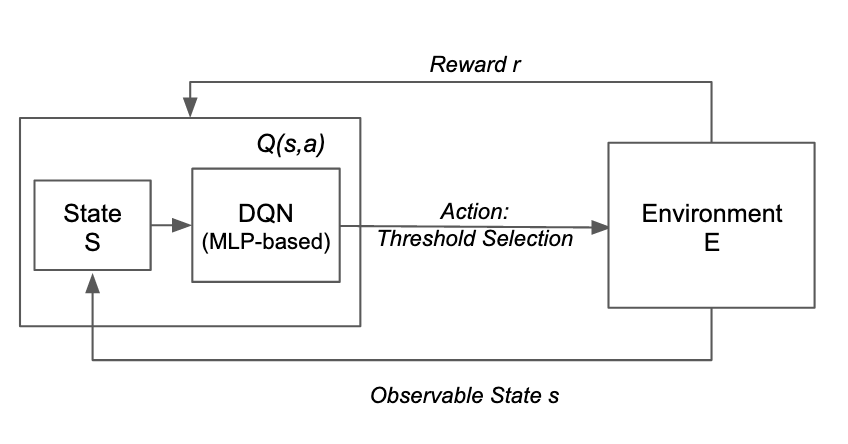}
\vspace{-5pt}
\caption{Deep Q-Network based adaptive threshold selection}
\label{fig:dqn}
\end{figure}

We then model the alert threshold selection problem as sequential decisions making: 
Given environment $E$, the threshold selection agent has to select one threshold at each hour, $a_{t}$ where $a_{t}\in A$ and $A$ has all $K$ thresholds to interact with $E$. At time $t$, the agent is in state $s_{t}$ and takes action $a_{t}$. Then its state turns into $s_{t+1}$ with reward $r_{t}$ obtained. During the day, the agent will have to interact with environment 24 times i.e. select one threshold for each hour of the day. In this sequential decision making task, the state of the agent $s_{t+1}$ only depends on the previous state $s_{t}$ and the action $a_{t}$. Each action will return an immediate reward and the goal is to select the actions by following a policy that maximizes the future rewards. 

The state of the agent can be any descriptive attribute of the agent (e.g. hour of the day and the monetary value of the fraud loss), as the internal states are only partially observable. More importantly, since the agent is expected to optimize the alert distribution under processing capacity constraints, the cumulative consumption of the daily processing capacity (created by each action) will be used as one of the agents state elements. When this capacity hits its limit $C$, the rest of the transactions for the day will be treated as non-fraudulent transactions. This formalism gives rise to a large but finite Markov Decision Process (MDP), for which we can apply reinforcement learning methods. By simply observing the sequences of $s_1,a_1,s_2,a_2,...,s_t,a_t$, a favorable threshold selection strategy can be learned for each hour. Fig. \ref{fig:dqn} illustrates the formalism of this sequential decision making process.

\subsubsection{Deep Q-Learning Formulation}
For any finite Markov decision process, Q-learning finds an optimal policy by maximizing the expected value of the total reward over any and all successive steps, starting from the current state \cite{MnihKSGAWR13}. In Q-learning, actions not only return immediate rewards but also impact the future decisions, which aligns well with the nature of the alerting process. Therefore, at each time point $t$, a reward is actually discounted by a factor of $\gamma$, which offers a trade-off between immediate and long term rewards. The agent seeks to maximize the expected discounted return denoted as:
\begin{equation}
R_t=\sum^{\infty}_{i=t}\gamma^{i-t}r_{i}   
\end{equation}
where $\gamma\in[0,1]$. 
Given a deterministic policy $\pi$, the values of the state-action pair $(s,a)$, also known as Q-function \cite{SuttonB2018}, is defined as: 
\begin{equation}
Q^{\pi}(s,a) = \mathbb{E}\left[R_t|s_t=s,a_t=a,\pi\right]
\end{equation}
Preceding Q-function can be computed recursively with dynamic programming:
\begin{equation}
Q^{\pi}(s,a) = \mathbb{E}_{s'}\left[r+\gamma~\mathbb{E}_{a'\sim\pi(s')}\left[Q^{\pi}(s',a')\right]|s,a,\pi\right]
\end{equation}
With the optimal $Q^{*}(s,a)=\max_{\pi}Q^{\pi}(s,a)$, optimal Q-function also satisfies the Bellman equation \cite{SuttonB2018}:
\begin{equation}
Q^{*}(s,a) = \mathbb{E}_{s'}\left[r+\gamma~\underset{a'}{\max} Q^{*}(s',a')|s,a\right]
\end{equation}
As a part of the model-free reinforcement learning approach, Q-learning does not require a model of the environment. It can handle problems with stochastic transitions and rewards without requiring adaptations. In the reinforcement learning community, it is common to use a linear function to approximate the optimal Q-function, $Q^{*}(s,a)$. However, instead of linear approximators, neural networks (Q-networks) can also be used thanks to their powerful function approximation capabilities. A Q-network parameterized by $\theta$ can be trained by minimizing a sequence of loss functions $L_i(\theta_i)$ that changes at each iteration $i$:
\begin{equation}
L_i(\theta_i) = \mathbb{E}_{s,a,r,s'}\left[\left(r+\gamma~\underset{a'}{\max} Q(s',a';\theta_{i-1})-Q(s,a;\theta_i)\right)^{2}\right]
\end{equation}

In this work, we use a DQN to better model the continuous states and the complex data distributions in the underlying problem. It was also selected for its relative simplicity, which is well suited in a practical use case such as alert generation.  Due to the low dimensionality and complexity of the alert use case compared with traditional gaming applications, DQN provides an efficient and practical option to complex alternatives \cite{WangFL2015,LillicrapHPH2015,MnihBMG2016,EspeholtSMS2018}.

\subsection{Deep Q-Network-based Adaptive Alert Threshold Selection}
In the proposed approach the agent has to take an action every hour, based on the acquired information about the state, action and reward for the previous hour. Its goal is to eventually maximize the final return at the end of one episode, where overly aggressive and overly conservative policies yield sub-optimal results. An aggressive agent tends to use up all the available capacity to maximize its return at the beginning. On the other hand, a conservative agent saves capacity with the cost of missing actual fraud transactions. The goal of the proposed approach is to find a balanced policy to avoid such scenarios. The setup for state, action and reward are given in detail as follows:

\textbf{\textit{State:}} Although the environment is only partially observable, we use 5 descriptive attributes to construct the state set for each hour: \begin{enumerate}[topsep=0pt]
\item{\textit{H} - Hour of the day} 
\item{\textit{S} - Total transaction amount for all confirmed fraud}
\item{\textit{L} - Total transaction amount for all missed fraud}
\item{\textit{CC} - Utilized processing capacity, which measures the processing capacity constraints for the system indirectly. As the system will drop all alerts after hitting the maximum capacity, the proposed algorithm can leverage this information to better balance the hourly thresholds}
\item{\textit{T} - Score Threshold}
\end{enumerate}
These attributes provide quantitative measures to guide the agent and derive a feasible policy to minimize fraud losses under the given constraints.

\textbf{\textit{Action:}} $K$ possible score thresholds, determined by the score distribution of upstream machine learning score engine. By selecting one of the thresholds as the action, the system treats any transaction with higher score as potential fraud and an alert will be generated. In practice, it is possible to have hundreds of score bands generated by the score engine, which creates a huge action space and hence poses a challenge on RL training. During the simulation period, each individual threshold shows significant performance fluctuations from time to time. We reduce the action space by selecting a subset of the thresholds. This accelerates the reinforcement learning training.

\textbf{\textit{Reward:}} We use  $(S-L)*H$ as the reward for each hour. This variable takes into account both (i) the actual fraud savings from the detected fraud transactions, (ii) the fraud losses caused by missing the fraud. By adding $H$ into the reward calculation, we seek to balance the strategy over time to prevent over-emphasizing early hours of the day.
In this paper we use a Multi-Layer Perceptron (MLP) to approximate the Q-function to better fit the alert use case, which is slightly different from the original implementation which uses convolutional neural networks \cite{MnihKSGAWR13,MinhKSR2015}.

\cite{MnihKSGAWR13} illustrates that learning directly from consecutive samples is inefficient due to the strong correlations between the samples. Randomizing the samples breaks these correlations and therefore reduces the variance of the updates. 
To improve the training efficiency, we use \textit{Experience Replay} \cite{MnihKSGAWR13}. The agents experiences at each time point, $e_t =(s_t,a_t,r_t,s_{t+1})$ is accumulated into a \textit{Replay Memory} $D_t =\{e_1,e_2,...,e_t\}$. While training the Q-network, a uniformly sampled mini-batch of experiences from $D$ is used instead of the current experience. In practice, the size of \textit{replay} memory can be predefined. The oldest experiences are removed when new experiences are acquired. This improves the data efficiency by reusing the experiences seen in the past and lowers the variance as randomly sampled experiences have less correlation in each update. For the hourly adaptive threshold selection, this technique helps avoid selecting the same threshold over consecutive hours. The loss sequence can be modified into its mini-batch version as follows:
\begin{equation}
\label{eq:final_cost}
L_i(\theta_i) = \mathbb{E}_{(s,a,r,s')\in U(D)}\left[\left(r+\gamma~\underset{a'}{\max} Q(s',a';\theta_{i-1})-Q(s,a;\theta_i)\right)^{2}\right]
\end{equation}
where $U(D)$ refers to uniformly sampled mini-batch from \textit{replay} memory $D$.

\begin{algorithm}[ht!]
\SetAlgoLined
Initialize replay memory \textit{D} to capacity \textit{N}\;
Initialize Q-network with random weights\;
\While{total number of training iterations not reached}{
    \For{Day from 1 to Max}{
        Reset consumed capacity $CC=0$, hour $H=1$\;
        \For{Hour ($t$) from 1 to 24}{
            With probability $\epsilon$ select a random action (threshold) $a_t$\;
            Otherwise, select $a_t = \underset{a}{\max} Q^{*}(s_{t-1},a_{t-1};\theta)$\;
            Execute action $a_t$ and calculate reward $r_t$\;
            Update state $s_t$\;
            Store $(s_{t-1},a_t,r_t,s_t)$ into replay buffer \textit{D}\;
            Sample random mini-batch of transitions $(s_{t-1},a_t,r_t,s_t)$ from \textit{D}\;
            Calculate $r+\gamma~\underset{a}{\max} Q(s_t,a_t;\theta)$ for the mini-batch\;
            Perform a gradient descent step and update Q-network to optimize Eq. \ref{eq:final_cost}.
        }
    }
 }
\caption{Deep Q-Network-based Adaptive Threshold Selection Algorithm for Fraud Alert Generation}
\label{algo:algorithm}
\end{algorithm}

To ensure adequate exploration of the state space, we adopt another DQN technique to select a random action with probability $\epsilon$, while following DQN output at $1-\epsilon$. This exploration allows the proposed method to search for optimal threshold selection policy in different conditions (hour of the day ($H$) and consumed processing capacity ($CC$)). The detailed algorithm of the proposed approach is presented as pseudo-code in Algorithm \ref{algo:algorithm}.

\section{Experimental Analysis}
\label{sec:simulation}
\subsection{Simulation Setup}
For the experimental analysis we constructed a two stage simulation set up with a fraud scoring stage followed by Deep Q-Learning-based threshold selection. For fraud scoring, an XGBoost (XGB) model was selected due to its performance after experimenting with XGB and Multi-Layer Perceptron (MLP)\footnote{Note this is not the MLP used in Deep Q-Network} architectures. The hyper-parameters and architectures of these models are shown in the Appendix. XGB was configured to output a fraud score range of 1-99.

In the threshold selection stage, a 3-layer MLP was used to build the Deep Q-learning model. A dataset with 724K historical credit card transactions was used for experimental analysis. The dataset includes transactions from January 2016 to December 2016 with $1.63\%$ fraud rate. No further sampling was performed. The reversed and multi-swipe credit transactions were removed to avoid data duplication. Table \ref{tab:dataset} shows the number of transactions and the total transaction amounts for fraud and non-fraud populations for each month. In practice, static thresholds are typically selected during the model retraining periods with few years intervals. In the experimental set up, due to the length of the dataset, static thresholds are selected and optimized using recent data. This gives the static thresholds an optimistic performance outlook, which should be interpreted as an upper bound for the production alert systems. 

\begin{table}[ht!]
\caption{Overview of the transaction and fraud distribution in the experimental dataset.}
\label{tab:dataset}
\scalebox{0.8}{
\begin{tabular}{lrrrr}
\toprule
\multicolumn{1}{c}{Month} & \multicolumn{1}{c}{Nonfraud \#} & \multicolumn{1}{c}{Fraud \#} & \multicolumn{1}{c}{Nonfraud \$} & \multicolumn{1}{c}{Fraud \$} \\ \hline
Jan                       & 56,881                           & 1027                         & 8,364,615.62                      & 238,007.90                     \\ 
Feb                       & 54,137                           & 963                          & 7,878,265.58                      & 219,051.49                    \\ 
Mar                       & 58,695                           & 1014                         & 8,483,692.67                      & 239,164.15                    \\ 
Apr                       & 57,516                           & 961                          & 8,208,244.23                      & 218,509.09                    \\ 
May                       & 60,342                           & 1031                         & 8,528,550.70                       & 244,639.65                    \\ 
Jun                       & 59,551                           & 985                          & 8,331,551.11                      & 214,606.98                    \\ 
Jul                       & 61,847                           & 998                          & 8,537,641.36                      & 224,170.75                    \\ 
Aug                       & 62,766                           & 971                          & 8,555,874.67                      & 225,804.07                    \\ 
Sep                       & 61,704                           & 958                          & 8,374,086.77                      & 209,563.76                    \\ 
Oct                       & 64,332                           & 998                          & 8,699,876.34                      & 226349.65                    \\ 
Nov                       & 62,884                           & 904                          & 8,442,234.66                      & 196,508.88                    \\ 
Dec                       & 63,887                           & 970                          & 8,397,197.87                      & 218,243.68                    \\ 
\hline
Total	                  & 724,542	                        & 11780	
& 10,0801,831.58	                  & 267,4620.05                   \\
\bottomrule
\end{tabular}
}
\end{table}
All static thresholds were analyzed over the January-March period. The highest performing thresholds of Threshold.0 - Threshold.10 (with corresponding scores of 56-66) were pre-selected to build the action space over the simulation period (for which the performance rankings are different). DQN-based threshold selection was designed to select a threshold out of Threshold.0 - Threshold.10 every hour. Transactions from March to September were used for DQN training.  October - December was used to test the DQN based adaptive threshold policy. One episode is set to one day and thus the entire session consists of 214 days. The trained DQN model was run from Day 1 of the testing period. 

\subsection{Performance Metrics}
Most fraud detection models optimize transaction-level metrics such as cross-entropy. In this study, the connections between the fraud savings and losses are made with the following outcomes: 
\begin{itemize}[leftmargin=1em]
\item \textit{True Positive:}  Fraudulent transaction was identified successfully, hence the monetary value of the transaction is considered \textit{fraud saving}.
\item \textit{True Negative:}  Non-fraudulent transaction was predicted as non-fraud. Most transactions fall into this category, as fraud frequencies are relatively small. This category does not impact model performance in any regard.
\item \textit{False Negative:} Fraudulent transaction was  identified incorrectly as non-fraud. This category represents the fraud loss.
\item \textit{False Positive:} Non-fraudulent transaction was incorrectly identified as fraud. Typically such labeling is associated with declined transactions, which does not directly yield any direct fraud savings or losses. In this study, we decided to not include this category.
\end{itemize}
For the experimental analysis, Net Fraud Savings is defined as the difference between Total Fraud Savings and Losses and \textit{Cumulative Net Fraud Savings (CNFS)} is the sum of Net Fraud Savings up to that point in time.  We use the \textit{Cumulative Net Fraud Savings} as the primary financial performance metric.  In an alert processing system, the performance of each threshold can only be evaluated after the alert is resolved, which can happen over a wide range, from a few seconds after the alert, to weeks after the transaction through a fraud claim submission. For the experimental analysis, we use the available ground truth for each transaction. We assume that $90\%$ of the alerts receive responses within an hour, and $10\%$ of non-alerted fraud was reported through the fraud claim submission process. Each hour resolved alerts can be used to calculate the reward function. However, the performance metrics are calculated later, when the ground truth for all alerts becomes available.

Similarly, the overall efficiency of the alert processing system is of great importance. As discussed earlier, an alert processing system may be (i) \textit{over-alerting:} during which too many alerts are generated, and alerts get dropped as they exceed the capacity. Furthermore, customer satisfaction degrades during over-alerting as excessive number of alerts are sent to the customers. (ii) \textit{under-alerting:} during which the alert system does not produce sufficient number of alerts and many fraud cases are undetected. This also degrades customer satisfaction as the customers are not alerted for fraud  transactions. To capture these operational and customer satisfaction effects, we use the \textit{Total Number of Over-Alerts and Under-Alerts} as a secondary performance metric. 

\subsection{Deep Q-Network Architecture}

The proposed method uses a three-layer MLP with $\{20,10,11\}$ neurons for each layer. Input to the MLP is the state vector for each time point. All features in the state vector are normalized to $\left[0,1\right]$. The first 2 layers use RELU. The last layer uses a linear activation function since it represents the Q-value as the output of the neural network. Adam \cite{Adam} optimizer with a learning rate of $0.0001$ was used over the Mean Squared Error-based loss function. 

For the reinforcement learning setup, the reward discount $\gamma$ is set to $0.9$.  Exploration probability $\epsilon$ follows a schedule such that it starts from $0.5$ and reduces by $5\%$ per iteration until $0.1$. This exploration schedule is designed to fully explore state space in early stage of the training and gradually switch to exploitation mode when approaching the convergence. The experience replay buffer size of $N$ and mini-batch size of $M$ are set to $160,000$ and $1024$ empirically. Lastly, we set the total number of iterations for the training and daily alert capacity to $100$ and $C_{max}=500$ respectively (based on experiments with different values).


\subsection{Experimental Results}

\begin{table}[ht!]
\caption{Highest performing static threshold based on \textit{CNFS} over March - September.}
\label{tab:best_perf_month}
\scalebox{0.9}{
\begin{tabular}{lrrrrrrrr}
\toprule
\multicolumn{1}{c}{Month} & \multicolumn{1}{c}{Mar} & \multicolumn{1}{c}{Apr} & \multicolumn{1}{c}{May} & \multicolumn{1}{c}{Jun} & \multicolumn{1}{c}{Jul} & \multicolumn{1}{c}{Aug} & \multicolumn{1}{c}{Sep}\\ 
\midrule
Threshold & Thr 7   & Thr 7 & Thr 5      & Thr 6        &Thr 5          & Thr 5     &Thr 4        \\
\bottomrule
\end{tabular}
}
\end{table}


\subsubsection{Static threshold performance fluctuations over daily and monthly periods} 
\begin{figure}[ht!]
\centering
\includegraphics[width=0.85\linewidth]{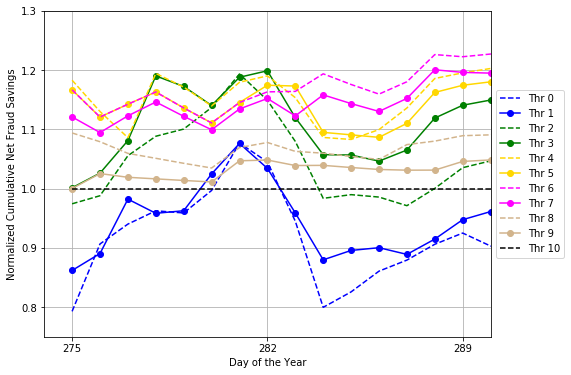}
\vspace{-5pt}
\caption{Normalized Cumulative Net Fraud Savings comparison for first two weeks in October}
\label{fig:case_study}
\end{figure}

\hfill\\
Static thresholds frequently experience performance fluctuations over time. As discussed in the earlier sections this causes over and under-alerting periods, affecting the fraud losses as well as the overall efficiency of the alert systems. Table \ref{tab:best_perf_month} shows the highest performing static threshold for March-September period, where no static threshold consistently outperforms others to warrant a clear selection. 
Similarly, Fig. \ref{fig:case_study} shows a more detailed view over a 2 week period from the dataset. The performance fluctuations can be seen through the Cumulative Net Fraud Savings (normalized over Threshold.10). Static thresholds frequently crossover, where no single static threshold consistently outperforming others. 

Both daily and monthly comparisons highlight the difficulties of selecting static thresholds in a dynamic environment such as fraud detection. 

\subsubsection{The operation of the adaptive threshold selection during the day}
\begin{figure}[ht!]
\centering
\includegraphics[keepaspectratio,width=0.95\columnwidth]{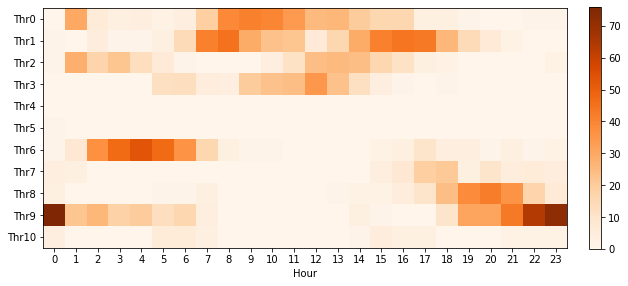}
\setlength{\abovecaptionskip}{-2pt}
\setlength{\belowcaptionskip}{-8pt}
\caption{Average threshold selection by the adaptive threshold policy for each hour over October-December.}
\label{fig:policy_sample}
\end{figure}

\hfill\\
Fig. \ref{fig:policy_sample} shows the distribution of the thresholds selected by the adaptive threshold selection policy for every hour of the testing period (the darker colors indicate a higher chance of being selected by DQN). This illustrates how the learning agent makes decisions and selects thresholds over a 24-hour period.  Despite showing strong performance as static thresholds,  Threshold.4 and Threshold.5 were very infrequently selected by the adaptive threshold selection policy. Instead, the dynamic policy utilizes other thresholds such as Threshold.9 and Threshold.0, Threshold.1, which were more frequently chosen to adapt dynamically to the underlying changes. At the beginning of the day the dynamic policy tends to adopt a more conservative strategy, by using higher thresholds to avoid over-alerting. Due to the transaction volume variations, it utilizes more aggressive thresholds during the business hours. Later in the day, it switches back to more conservative thresholds and produces fewer alerts to stay within the constraints of the daily alert processing capacity. Intuitively, this learned policy shares quite some similarities with how a human specialist would balance the performance and capacity constraints. 

\subsubsection{Fraud savings for static thresholds and deep Q-learning based adaptive threshold selection}
\hfill\\
Table \ref{tab:performance} shows Cumulative Net Fraud Savings for the static thresholds and the dynamic threshold selection over October-December. Monthly values were calculated at the end of each month.  
On average, mid-size and large banks process 1-10 million and 10-30+ million credit transactions per day respectively. As a result, the cumulative net fraud savings values for the proposed approach are expected to scale proportionally to orders of magnitude higher values in deployment systems. As the absolute difference of CNFS is highly dependent on the volume of transactions, Table \ref{tab:performance} also provides the relative performance normalized to the worst performing threshold, Threshold.10.  

\begin{table}[ht!]
\caption{Performance comparison of the Deep Q-learning-based Adaptive Threshold Selection Policy against the static baseline thresholds over October-December.                                                         }
\label{tab:performance}
\scalebox{0.75}{
\begin{tabular}{|l|r|r|r|r|r|r|}
\hline
          & \multicolumn{3}{c|}{Cumulative Net Fraud Savings (\$)}                         & \multicolumn{3}{c|}{Relative Improvements}                                     \\ \hline
 & \multicolumn{1}{c|}{Oct} & \multicolumn{1}{c|}{Nov} & \multicolumn{1}{c|}{Dec} & \multicolumn{1}{c|}{Oct} & \multicolumn{1}{c|}{Nov} & \multicolumn{1}{c|}{Dec} \\ \hline
Thr0   & 91,627.65                 & 172,916.59                & 272,233.33                & 12.10\%                  & 2.18\%                   & 6.28\%                   \\ \hline
Thr1    & 94,934.01                 & 179,207.39                & 285,610.45                & 16.15\%                  & 5.90\%                   & 11.50\%                  \\ \hline
Thr2    & 100,236.95                & 195,215.73                & 304,032.85                & 22.64\%                  & 15.36\%                  & 18.69\%                  \\ \hline
Thr3    & 105,120.61                & 203,428.47                & 312,060.59                & 28.61\%                  & 20.22\%                  & 21.82\%                  \\ \hline
Thr4    & 105,916.35                & 209,866.51                & 317,246.51                & 29.59\%                  & 24.02\%                  & 23.85\%                  \\ \hline
Thr5   & 104,516.99                & 207,827.61                & 317,227.83                & 27.87\%                  & 22.82\%                  & 23.84\%                  \\ \hline
Thr6    & 106,207.49                & 207,405.47                & 311,000.19                & 29.94\%                  & 22.57\%                  & 21.41\%                  \\ \hline
Thr7    & 100,229.41                & 196,630.73                & 296,710.21                & 22.63\%                  & 16.20\%                  & 15.83\%                  \\ \hline
Thr8    & 93,202.05                 & 185,642.93                & 279,583.21                & 14.03\%                  & 9.71\%                   & 9.15\%                   \\ \hline
Thr9    & 86,321.93                 & 177,035.09                & 268,263.65                & 5.61\%                   & 4.62\%                   & 4.73\%                   \\ \hline
Thr10    & 81,733.93                 & 169,219.55                & 256,156.91                & -                   & -                   & -                  \\ \hline
Ours    & \textbf{112,058.71}                & \textbf{217,673.93}                & \textbf{334,158.41}                & \textbf{37.10\%}                  & \textbf{28.63\%}                  & \textbf{30.45\%}                  \\ \hline
\end{tabular}
}
\end{table}

For October, Threshold.6 performed the best against other static thresholds, while Threshold.4 was the highest performing for November and December. At the end of December, Threshold.5 and Threshold.4 perform very close to each other.

Overall, Deep Q-learning based adaptive threshold selection technique outperforms the highest performing static thresholds of each month by $6\%$ (on average). For the October-December period, it provides $37.10\%$, $28.63\%$ and $30.45\%$ improvement over Threshold.10 for each month.






\subsubsection{Customer satisfaction and operational comparison through over-alert \& under-alert counts}
\hfill\\
As discussed in the earlier sections, in a production system relying purely on a single performance metric is not sufficient to assess the overall system improvement.  
Table \ref{tab:over_under_alert} shows the number of over and under-alerts for the highest performing static thresholds (Threshold.4-Threshold.6) and the adaptive threshold selection policy. Over and under-alerting impact the operational efficiency of the system and serve as a metric for customer satisfaction. 

Compared to the highest performing static thresholds, the proposed adaptive threshold selection approach yields the lowest number of over and under-alerts. Over the 3 month period, it provides 9.6\% improvement over the best static threshold for each month. 

\begin{table}[ht!]
\caption{Total number of over-alert and under-alerts for the static thresholds vs. dynamic threshold selection over October-December.}
\label{tab:over_under_alert}
\scalebox{0.8}{
\begin{tabular}{lrrrr}
\toprule
Month & Threshold.4 & Threshold.5 & Threshold.6 & Ours \\ 
\midrule
Oct & 727   & 946   & 1353  & \textbf{607}  \\
Nov & 1350  & 1861  & 2715  & \textbf{1284} \\
Dec & 2135  & 2933  & 4266  & \textbf{1974} \\
\bottomrule
\end{tabular}
}
\end{table}

\section{Conclusions}
Alert systems are pervasively deployed across all payment channels in retail banking. In an alert system, threshold selection process determines the transactions to alert based on the fraud scores. For simplicity current alert systems use static thresholds, which are not able to optimize alert generation for alert processing capacity constraints. In this study, we propose a Deep Q-Network based reinforcement learning approach to improve the alert threshold selection in fraud alert systems. Experimental results show that the resulting trained agent outperforms static threshold policies both in terms of net fraud saving and metrics for customer service. Furthermore, this learned policy can be used to better adapt to the changes in the environment automatically, including emerging transaction and fraud patterns.

\bibliography{bib}
\bibliographystyle{ACM-Reference-Format}

\section*{Appendix}


\begin{table}[h!]
\caption{\small{Hyper-parameters/Architecture of XGBoost\& MLP}}
\label{tab:xg_mlp}
\scalebox{0.9}{
\begin{tabularx}{0.483\textwidth}{lX}
\toprule
Model  & Hyper-parameters/Architecture  \\
\midrule
XGBoost & colsample\_bytree=0.8, gamma=0.9, max\_depth=3, min\_child\_weight=2.89, reg\_alpha=3, reg\_lambda=40, subsample=0.94, learning\_rate=0.1, n\_estimators=100 \\
\midrule
MLP & Dense(100/RELU)-BatchNorm-Dropout(0.5)-Dense(50/RELU)-Dropout(0.5)-Softmax(2),learning\_rate=0.001,batch\_size=512\\
\bottomrule
\end{tabularx}
}
\end{table}
\end{document}